\documentclass{article}

\usepackage[final]{nips_2017}

\usepackage[utf8]{inputenc} 
\usepackage[T1]{fontenc}    
\usepackage{hyperref}       
\usepackage{url}            
\usepackage{booktabs}       
\usepackage{amsfonts}       
\usepackage{nicefrac}       
\usepackage{microtype}      
\usepackage{siunitx}

\usepackage{graphicx}

\usepackage[ruled]{algorithm2e}

\usepackage{color}

\title{Enhanced Experience Replay Generation for Efficient Reinforcement Learning}

\author{
  Vincent Huang* \\
  Ericsson AB\\
  Stockholm, Sweden \\
  \texttt{vincent.a.huang@ericsson.com} \\
  \And
  Tobias Ley* \\
  Ericsson AB\\
  Stockholm, Sweden \\
  \texttt{tobias.ley@ericsson.com} \\
  \And
  Martha Vlachou-Konchylaki* \\
  Ericsson AB\\
  Stockholm, Sweden \\
  \texttt{martha.vlachou-konchylaki@ericsson.com} \\
  \And
  Wenfeng Hu* \\
  Ericsson AB \\
  Stockholm, Sweden \\
  \texttt{wenfeng.hu@ericsson.com} \\
}

\begin{document}

\maketitle

\begin{abstract}
Applying deep reinforcement learning (RL) on real systems suffers from slow data sampling. We propose an enhanced generative adversarial network (EGAN) to initialize an RL agent in order to achieve faster learning. The EGAN utilizes the relation between states and actions to enhance the quality of data samples generated by a GAN. Pre-training the agent with the EGAN shows a steeper learning curve with a 20\% improvement of training time in the beginning of learning, compared to no pre-training, and an improvement compared to training with GAN by about 5\% with smaller variations. For real time systems with sparse and slow data sampling the EGAN could be used to speed up the early phases of the training process.

\end{abstract}

\section{Introduction}

In 5G telecom systems, network functions need to fulfill new network characteristic requirements, such as ultra-low latency, high robustness, quick response to changed capacity needs, and dynamic allocation of functionality. With the rise of cloud computing and data centers, more and more network functions will be virtualized and moved into the cloud. 
Self-optimized and self-care dynamic systems with fast and efficient scaling, workload optimization, as well as new functionality like self-healing, parameter free and zero-touch systems will assure SLA (Service Level Agreements) and reduce TCO (Total Cost of Ownership). Reinforcement learning, where an agent learns how to act optimally given the system state information and a reward function, is a promising technology to solve such an optimization problem. 
 
Reinforcement learning is a technology to develop self-learning SW agents, which can learn and optimize a policy based on observed states of the environment and a reward system. An agent receives observations from the environment in state $S$ and selects an action to maximize the expected future reward $R$. Based on the expected future rewards, a value function $V$ for each state can be calculated, and an optimal policy $\pi$ that maximizes the long term value function can be derived. In a model-free environment, the RL agent needs to balance exploitation with exploration. Exploitation is the strategy to select actions based on previously learned policy, while exploration is a strategy to search for better policies using actions not from the learned policy. Exploration creates opportunities, but also induces the risk that choices done during this phase will not generate increased reward. 

In real-time service-critical systems, exploration can have an impact on the service quality. In addition, sparse and slow data sampling, and extended training duration put extra requirements on the training phase. This paper proposes a new approach for pre-training the agent based on enhanced GAN data sampling to shorten the training phase, to address the training limitation options of environments with sparse and slow data sampling.

The paper is organized as follows. In Section \ref{sec_back}, we give a brief overview of reinforcement learning, generative adversarial networks and their recent development. In Section \ref{sec_egan}, we present our proposed approach of a pre-training system with enhanced GAN. Experiment results are presented in Section \ref{sec_res}. Finally, we give concluding remarks and discussions in Section \ref{sec_con}.


\section{Background}
\label{sec_back}

\subsection{Reinforcement Learning}

Reinforcement learning is generally the problem of learning to make decisions by maximizing a numerical reward signal. \citep{Sutton:1998}. A reinforcement learning agent receives an observation $o_t$ from the environment it interacts in state $s_t$, and selects an action $a_t$ so as to maximize the total expected discounted reward $G_t$. The action, drawn from the action space $A$, is calculated by a policy $\pi(a_t|s_t)$. Every time the policy is executed, a scalar reward $R_s^a$ is returned from the environment, and the agent transitions to the next state, $s_{t+1}$, following the state transition probability $P_{ss'}^a = P(s'| s, a)$. 

We can define the state value function $V^{\pi}(s)$ as the expected return at state $s$, following policy $\pi$, and the action value function $Q^{\pi}(s)$ as the expected return taking action $a$, while in state $s$, following policy $\pi$.

The reinforcement learning agent tries to maximize the expected return by maximizing the value function $V^{\pi}$ :
\begin{equation}
V^{\pi}(s) = \Sigma_{a \in A}\pi(a|s)(R_s^a+\gamma \Sigma_{s' \in S}P_{ss'}^aV^{\pi}(s'))
\end{equation}

An approach of maximizing $V^{\pi}(s)$ is using policy gradients (PG), in which the policy is parametrized and optimized by calculating the gradients using supervised learning, while iteratively adjusting the weights by backpropagating the gradients into the neural network.

Most reinforcement learning work uses simulated environments like OpenAI Gym \citep{openaigym} and can achieve good results by running many episodes \citep{duan2016benchmarking, mnih2015human}. Compared to simulated environments, real environments have different characteristics and different training strategies need to be applied. The agent has access only to partial, local information, which can be formalized as a Decentralized Partial-Observable Markov Decision Process (Dec-POMDPs) \citep{Oliehoek2012}. Further, it is either not possible or too expensive to do exhaustive exploration strategies in a real production system, which might cause service impact. Finally, sparse data, low data sampling rate, and slow reaction time to actions greatly limit the possibility to train an agent in an acceptable time frame \citep{duan2016benchmarking}. New, sample efficient algorithms such as Q-Prop \citep{gu2016q} have been proposed, that provide substantial gains in sample efficiency over trust region policy optimization (TRPO) \citep{schulman2015trust}. Methods such as actor critic algorithms \citep{mnih2016asynchronous}, as well as combinations of on-policy and off policy algorithms \citep{ocombining} have been tested to beat the benchmarks. Other approaches using supervised learning have been also tested \citep{pinto2016supersizing}. Still, the need of increasing sample efficiency to speed-up training time is imperative in real production systems that only allow for sparse data sampling. 

\subsection{Generative Adversarial Networks}

A second trend in deep learning research has been generative models, especially Generative Adversarial Nets (GAN) \citep{gans}, and the connection to reinforcement learning \citep{finn2016connection,yu2017seqgan} has been discussed. GANs are used to synthesize data samples that can be used for training an RL agent. In our case, these synthesized data samples are used to pre-train a Reinforcement Learning Agent to speed-up the training time in the real production system. We will compare this method with different pre-training alternatives.

The essence behind Generative Adversarial Nets is an aversion between a generative model $G$, which learns the true data distribution, and a discriminative model $D$, which evaluates the probability of a sample coming from the true distribution, rather than having been generated by $G$. 

The generator, modeled as a multilayer perceptron, is given inputs $z$, sampled from a noise distribution $p_z$. The network $G(z;\theta_g)$ is trained to learn the mapping from $z \sim p_z(z)$ to $x \sim p_{\tt{data}}(x)$, where $p_{\tt{data}}$ is the true data distribution. The discriminator, $D(\bf{x};\theta_d)$, also represented by a multilayer perceptron, is given as input either the generated sample $x\sim p(x|z)$ or a true data point $x\sim p_{\tt{data}}$. $D(x)$ is learning the probability of $x$ originating from the true distribution. 

By training both models in parallel, we can converge to a single solution where $G$ can eventually capture the training data distribution, and $D$ cannot discriminate between true and generated samples. 


\section{Enhanced GAN}
\label{sec_egan}

The object of GAN can be considered as the minmax game. The discriminator tries to maximize a value function, while the generator tries to minimize it, as shown below.
\begin{equation}
\min_{G}\max_{D}V(D,G)
\end{equation}
where, the value function $V(D,G)$ can be expressed as:
\begin{equation}
V(D,G)=\mathbb{E}_{x\sim p_{data}(x)}[\log D(x)]+\mathbb{E}_{z\sim p_{z}(z)}[\log(1-D(G(z)))]
\end{equation}
In our case, the training data set is the collected state($s$)-\{action, reward\}($a$) pairs.
Thus, the training data can be subset to two parts:
\begin{equation}
x=[x_{1}, x_{2}]=[(s_t,a),(s_{t+1},r)]
\end{equation}
Correspondingly, the generated data also consist of two parts:
\begin{equation}
G(z)=[G_{1}(z), G_{2}(z)]=[(s_t^\prime, a^\prime),(s_{t+1}^\prime, r^\prime) ]
\end{equation}
where $s^\prime$ and $a^\prime$ are the generated state($s$)-\{action, reward\}($a$) pairs.
Since the new state and reward depend on the current state and the selected action, there are latent relations between $x_{1}$ and $x_{2}$. The mutual information $I$ between $X_{1}$ and $X_{2}$ can be expressed as two entropy terms:
\begin{equation}
I(X_2;X_1)=H(X_2)-H(X_2|X_1)
\end{equation}
where $X_1$ represents the $(s_t^\prime, a^\prime)$ pair and $X_2$ represents the $(s_{t+1}^\prime, r^\prime)$ pair. The $(s_{t+1}^\prime, r^\prime)$ pair is dependent of the $(s_t^\prime, a^\prime)$ pair, therefore $I(X_2;X_1)$ cannot be zero.
To utilize this information, we can generate better quality experience replay data. To achieve this, we use the Kullback–Leibler divergence from $Q$ to $P$, where $P$ is the distribution of the generated action values $G_2(z)$ and $Q$ represents the distribution of derived dependent $(s_{t+1}'', r'')$ pair generated from $G_1(z)$ using the mutual information $I(X_2;X_1)$.
\begin{equation}
D_{KL}(P||Q)=\sum_{i}P(i)\log\frac{P(i)}{Q(i)}
\end{equation}
$I(X_2;X_1)$ can be obtained by training from the real experience replay data.
Thus, we can update the value function of $V(D,G)$ as
\begin{equation}
V(D,G)=\mathbb{E}_{x\sim p_{data}(x)}[\log D(x)]+\mathbb{E}_{z\sim p_{z}(z)}[\log(1-D(G(z)))]+\lambda D_{KL}(P||Q)
\end{equation}
where $\lambda$ is just a weighting factor. The last term is a regularization term to force the GAN to follow the relation between state and action-reward pair. When the GAN improves, $G_1(z)$ and $G_2(z)$ will follow the relations in the real experience replay data and the $KL$-divergence will tend to zero. The goal of the generator network is also to minimize this term.

\begin{figure}[h]
  \centering
  \includegraphics[width=90mm]{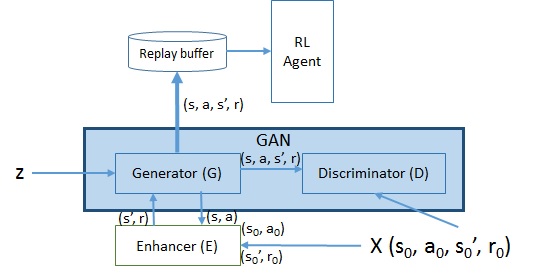}
  \caption{Enhanced GAN structure.}
  \label{fig_egan}
\end{figure}

The network architecture can be realized as in figure~\ref{fig_egan}.
Besides the normal GAN networks, an additional DNN has been added, to train the relations between state($s$) and \{action, reward\}($a$) pairs. The training procedure is shown in algorithm~\ref{alg_egan}.

\begin{algorithm}[H]
\caption{Data generation algorithm with EGAN}\label{alg_egan}
\SetAlgoLined
\KwData{Batch of quadruplets $D_r (s_t,a,s_{t+1},r)$ drawn from the real experience}
\KwResult{Unlimited experience replay samples $D_s (s_t,a,s_{t+1},r)$ which can be used for the pre-training of the reinforcement learning agent.}
\Begin{
 initialization\;
 \tcc{initializes the weights for generator and discriminator networks in GAN, as well as the enhancer network}
 training GAN\;
 \tcc{training a GAN network with the real experience data $D_r (s_t,a,s_{t+1},r)$}
 training enhancer\;
 \tcc{training an enhancer network with the real experience data $D_r (s_t,a,s_{t+1},r)$ to find the relations between $D_r (s_t,a)$ and $D_r (s_{t+1},r)$}
 \For{k iterations}{
  generate data $D_t (s_t,a,s_{t+1},r)$ with GAN\;
  \tcc{generate a test experience data set with GAN}
  improve GAN with enhancer\;
  \tcc{using the enhancer to calculate the discrepancy between $D_t (s_t,a)$ and $D_t (s_{t+1},r)$ and use this to update GAN}
 }
}
\end{algorithm}

In practice, we can update the GAN with the regularization term at the same time. However, it is also possible to update the regularization term separately. In a real system, where the data collection is slow, more training on the network can be performed while waiting fro inputs of the new experience replay data. We train the relation between the state($s$) and \{action, reward\}($a$) pairs whenever new experience data comes in. After we train the GAN with the normal settings, the network weights can be updated using the trained relations from the Enhancer.

Once the GAN has been trained, it is possible to generate unlimited experience replay data to pre-train the agent.


\section{Results}
\label{sec_res}
%

\begin{figure}[h]
	\centering
	\includegraphics[width=140mm]{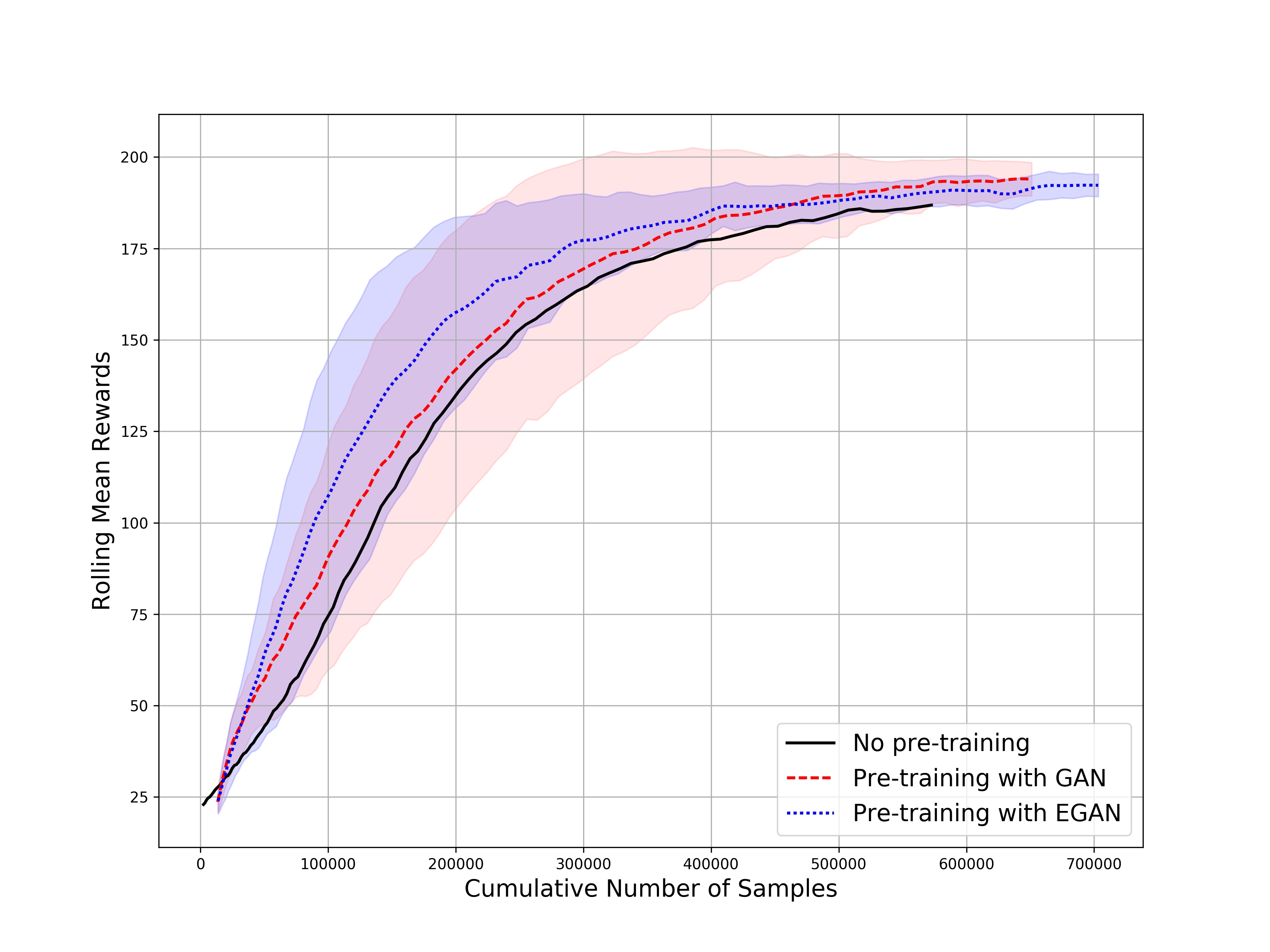}
	\caption{Comparison of with and without EGAN pre-training.}
	\label{fig_egan_performance}  
\end{figure}

We use the CartPole environment from OpenAI Gym to evaluate the EGAN performance, as shown in figure~\ref{fig_egan_performance}, with parameter settings listed in table~\ref{parameter_setting_table}. 
The figure shows the training of the PG agent after it has been pre-trained, therefore we observe a small offset of the pre-trained agents on the x-axis by around 10000 samples (500 episodes), while the agent with no pre-training starts at 0. 
The black solid line is the 100-episode rolling average reward over the total consumed samples of a PG agent, without any pre-training mechanisms. 
The red dash line and the blue dot line represent the performance of the PG agent with GAN and EGAN pre-training respectively. 
The EGAN uses 500-episode real experience, $D_r$, with randomly selected actions to train the GAN and Enhancer neural networks in the pre-training phase, and then generates 6000 batches of synthetic data, $D_t$, to update the policy network in the beginning of the training phase. 
For the no pre-training agent we set the total training episodes to 5500, to have a fair comparison with the EGAN over cumulative samples.

The samples for training the GAN and EGAN were collected using a random policy. 
Consequently, we expect a low initial performance for both pre-trained systems, but a more accurate value function estimation, thus a quicker learning curve since they are already initiated by generated samples. 
As a result, we can observe in figure \ref{fig_egan_performance} a faster increase of the reward for both agents pre-trained with GAN and EGAN. 
Both those networks can provide more modalities in the data space and since EGAN enhances the state-action-reward relation it can further improve the quality of the synthesized data, and the robustness of the system in terms of single standard deviation. 
We obtain, therefore, a 20\% higher sample efficiency for EGAN pre-training compared to no pre-training, and a 5\% improvement compare to pre-training with GAN without an enhancer. That means to reach a certain mean reward, less cumulative samples are needed, thus speeding up the training time.  


\begin{table}[t]
	\caption{EGAN simulation parameter settings}
	\label{parameter_setting_table}
	\centering
	\begin{tabular}{llll}
		\toprule
		\multicolumn{2}{c}{\textbf{Pre-training phase}}                             & \multicolumn{2}{c}{\textbf{Training phase}}               \\
		\midrule
		\emph{D and G network size}               & $[40, 20], [40, 20]$   & \emph{Policy network size}      & $[32]$ \\
		\emph{Enhancer network size}              & $[60, 60]$             & \emph{PG learning rate}      & $\num{1e-3}$ \\
		\emph{GAN learning rate}                  & $\num{5e-6}$           & \emph{PG discount factor}    & $0.99$ \\
		\emph{GAN sample size}                    & $64$                   & \emph{PG update frequency}    & $5$    \\
		\emph{Pre-training buffer size}           & $500$ episodes        & \emph{Training episodes}    & $5000$     \\
		\emph{EGAN pre-training iterations}       & $k=2$                  & \emph{Synthetic replay buffer size}   & $6000$   \\
		\bottomrule
	\end{tabular}
\end{table}

\begin{figure}[h]
	\centering
	\includegraphics[width=140mm]{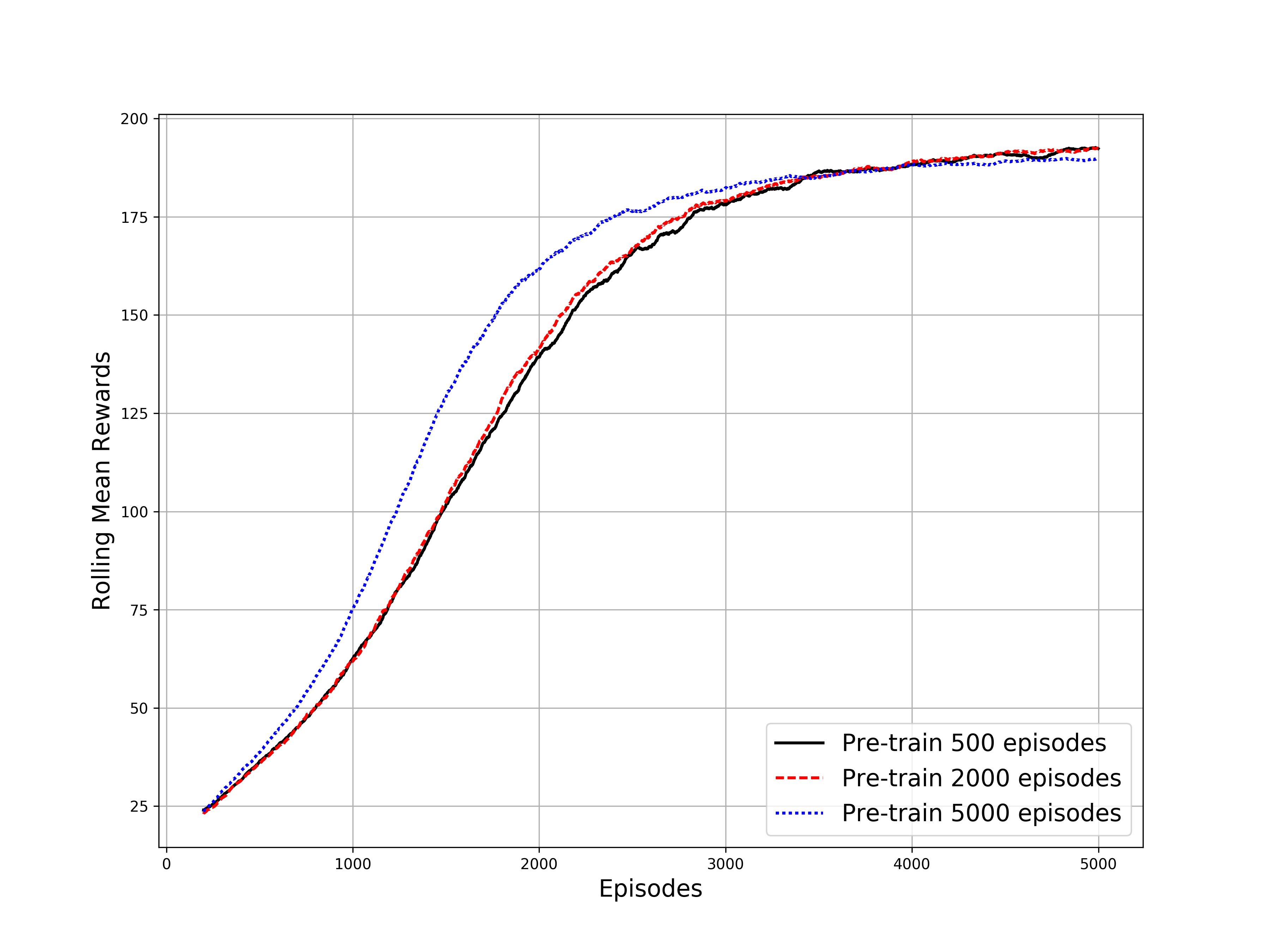}
	\caption{Comparison of different pre-training lengths.}
	\label{fig_egan_diff_length}  
\end{figure}

In order to test the hypothesis of bootstrapping the online training with DNN, GAN, and policy network pre-initialization, we trained our system with a varying number of pre-training lengths, demonstrating the results in figure~\ref{fig_egan_diff_length}.  The y axis represents again the 100-episode rolling average reward, while the x axis displays the online episode numbers. Figure  ~\ref{fig_egan_diff_length} demonstrates that  pre-training the generator networks with 5000 episodes results in a faster learning curve for the policy network. 

In a real production system, the pre-training could be achieved with saving prior data to pre-initialize the system, so as to aid it to converge faster, while also achieving a more stable training.
Therefore, it is of great importance to point out the fact that in real environments, where samples are expensive to produce, while also taking into consideration the episodes needed for the pre-initialization, pre-training the network with 500 episodes rather than 5000 is more efficient and cost-effective.

\section{Conclusions}
\label{sec_con}
In this work, we are tackling a fundamental problem of reinforcement learning applied to a real environment. The training normally takes long time and requires many samples. We first collected a small set of data samples from the environment, following a random policy, in order to train a GAN. The GAN is then used to generate unlimited synthesized data to pre-train an RL agent, so that the agent learns the basic characteristics of the environment. Using a GAN, we can cover larger variations of the random sampled data. We further improve the GAN with an enhancer, which utilizes the state-action relations in the experience replay data in order to improve the quality of the synthesized data. 

By using the enhanced structure (EGAN) we can achieve a 20\% faster than no pre-training, a 5\% faster learning than pre-training with a GAN, and a more robust system in terms of standard deviation. However, further work is needed to verify and fine-tune the system for achieving optimal performance.

Our next step is to explore and test this setup together with virtualized network functions in 5G telecom systems, where sample efficiency is crucial, and exploration can directly affect the service quality of the system. 



\bibliographystyle{unsrtnat}
\bibliography{ml-ref}

\begin{thebibliography}{13}
\providecommand{\natexlab}[1]{#1}
\providecommand{\url}[1]{\texttt{#1}}
\expandafter\ifx\csname urlstyle\endcsname\relax
  \providecommand{\doi}[1]{doi: #1}\else
  \providecommand{\doi}{doi: \begingroup \urlstyle{rm}\Url}\fi

\bibitem[Sutton and Barto(1998)]{Sutton:1998}
Richard~S. Sutton and Andrew~G. Barto.
\newblock \emph{Introduction to Reinforcement Learning}.
\newblock MIT Press, Cambridge, MA, USA, 1st edition, 1998.
\newblock ISBN 0262193981.

\bibitem[Brockman et~al.(2016)Brockman, Cheung, Pettersson, Schneider,
  Schulman, Tang, and Zaremba]{openaigym}
Greg Brockman, Vicki Cheung, Ludwig Pettersson, Jonas Schneider, John Schulman,
  Jie Tang, and Wojciech Zaremba.
\newblock Openai gym, 2016.

\bibitem[Duan et~al.(2016)Duan, Chen, Houthooft, Schulman, and
  Abbeel]{duan2016benchmarking}
Yan Duan, Xi~Chen, Rein Houthooft, John Schulman, and Pieter Abbeel.
\newblock Benchmarking deep reinforcement learning for continuous control.
\newblock In \emph{Proceedings of the 33rd International Conference on Machine
  Learning (ICML)}, 2016.

\bibitem[Mnih et~al.(2015)Mnih, Kavukcuoglu, Silver, Rusu, Veness, Bellemare,
  Graves, Riedmiller, Fidjeland, Ostrovski, et~al.]{mnih2015human}
Volodymyr Mnih, Koray Kavukcuoglu, David Silver, Andrei~A Rusu, Joel Veness,
  Marc~G Bellemare, Alex Graves, Martin Riedmiller, Andreas~K Fidjeland, Georg
  Ostrovski, et~al.
\newblock Human-level control through deep reinforcement learning.
\newblock \emph{Nature}, 518\penalty0 (7540):\penalty0 529--533, 2015.

\bibitem[Oliehoek(2012)]{Oliehoek2012}
Frans~A. Oliehoek.
\newblock \emph{Decentralized POMDPs}, pages 471--503.
\newblock Springer Berlin Heidelberg, Berlin, Heidelberg, 2012.
\newblock ISBN 978-3-642-27645-3.
\newblock \doi{10.1007/978-3-642-27645-3_15}.
\newblock URL \url{http://dx.doi.org/10.1007/978-3-642-27645-3_15}.

\bibitem[Gu et~al.(2016)Gu, Lillicrap, Ghahramani, Turner, and Levine]{gu2016q}
Shixiang Gu, Timothy Lillicrap, Zoubin Ghahramani, Richard~E Turner, and Sergey
  Levine.
\newblock Q-prop: Sample-efficient policy gradient with an off-policy critic.
\newblock \emph{arXiv preprint arXiv:1611.02247}, 2016.

\bibitem[Schulman et~al.(2015)Schulman, Levine, Abbeel, Jordan, and
  Moritz]{schulman2015trust}
John Schulman, Sergey Levine, Pieter Abbeel, Michael Jordan, and Philipp
  Moritz.
\newblock Trust region policy optimization.
\newblock In \emph{Proceedings of the 32nd International Conference on Machine
  Learning (ICML-15)}, pages 1889--1897, 2015.

\bibitem[Mnih et~al.(2016)Mnih, Badia, Mirza, Graves, Lillicrap, Harley,
  Silver, and Kavukcuoglu]{mnih2016asynchronous}
Volodymyr Mnih, Adria~Puigdomenech Badia, Mehdi Mirza, Alex Graves, Timothy
  Lillicrap, Tim Harley, David Silver, and Koray Kavukcuoglu.
\newblock Asynchronous methods for deep reinforcement learning.
\newblock In \emph{International Conference on Machine Learning}, pages
  1928--1937, 2016.

\bibitem[O’Donoghue et~al.()O’Donoghue, Munos, Kavukcuoglu, and
  Mnih]{ocombining}
Brendan O’Donoghue, R{\'e}mi Munos, Koray Kavukcuoglu, and Volodymyr Mnih.
\newblock Combining policy gradient and q-learning.

\bibitem[Pinto and Gupta(2016)]{pinto2016supersizing}
Lerrel Pinto and Abhinav Gupta.
\newblock Supersizing self-supervision: Learning to grasp from 50k tries and
  700 robot hours.
\newblock In \emph{Robotics and Automation (ICRA), 2016 IEEE International
  Conference on}, pages 3406--3413. IEEE, 2016.

\bibitem[{Goodfellow} et~al.(2014){Goodfellow}, {Pouget-Abadie}, {Mirza}, {Xu},
  {Warde-Farley}, {Ozair}, {Courville}, and {Bengio}]{gans}
I.~J. {Goodfellow}, J.~{Pouget-Abadie}, M.~{Mirza}, B.~{Xu}, D.~{Warde-Farley},
  S.~{Ozair}, A.~{Courville}, and Y.~{Bengio}.
\newblock {Generative Adversarial Networks}.
\newblock \emph{ArXiv e-prints}, June 2014.

\bibitem[Finn et~al.(2016)Finn, Christiano, Abbeel, and
  Levine]{finn2016connection}
Chelsea Finn, Paul Christiano, Pieter Abbeel, and Sergey Levine.
\newblock A connection between generative adversarial networks, inverse
  reinforcement learning, and energy-based models.
\newblock \emph{arXiv preprint arXiv:1611.03852}, 2016.

\bibitem[Yu et~al.(2017)Yu, Zhang, Wang, and Yu]{yu2017seqgan}
Lantao Yu, Weinan Zhang, Jun Wang, and Yong Yu.
\newblock Seqgan: sequence generative adversarial nets with policy gradient.
\newblock In \emph{Thirty-First AAAI Conference on Artificial Intelligence},
  2017.

\end{thebibliography}

\end{document}